%% file: BeyondFusion.tex
\documentclass[journal]{IEEEtran}
\usepackage{amsmath,amsfonts}
\usepackage{algorithmic}
\usepackage{array}
\usepackage[caption=false,font=normalsize,labelfont=sf,textfont=sf]{subfig}
\usepackage{textcomp}
\usepackage{stfloats}
\usepackage{url}
\usepackage{verbatim}
\usepackage{graphicx}
\usepackage{siunitx}
\usepackage{diagbox}
\usepackage{colortbl}
\usepackage{multirow}

\def\BibTeX{{\rm B\kern-.05em{\sc i\kern-.025em b}\kern-.08em
    T\kern-.1667em\lower.7ex\hbox{E}\kern-.125emX}}
\usepackage{balance}
\begin{document}
\title{BeyondFusion: Self-Aligned Latent Diffusion for Calibration-Free Infrared Super-Resolution and Infrared-Visible Fusion}

\author{
    \IEEEauthorblockN{
        Minchong Chen$^{1,\dagger}$,
        Xiaoyun Yuan$^{1,\dagger,*}$,~\IEEEmembership{Member,~IEEE},
        Minyu Cao$^{1}$,
        Jianing Zhang$^{2}$,\\
        Jun Zhang$^{2}$,
        Shuyang Liu$^{3,*}$,
        Xiaokang Yang$^{1}$,~\IEEEmembership{Fellow,~IEEE}
    }\\
    \IEEEauthorblockA{
        $^1$MoE Key Lab of Artificial Intelligence, Institute of AI, School of Computer Science, Shanghai Jiao Tong University, Shanghai, China\\
        $^2$Tsinghua University, Beijing, China\\
        $^3$Institute of Modern Optics, Nankai University, Tianjin, China\\
        Email: yuanxiaoyun@sjtu.edu.cn
    }
    \thanks{† These authors contributed equally to this work.}
    \thanks{* Corresponding author.}
}



\markboth{IEEE Transactions on Pattern Analysis and Machine Intelligence}%
{Chen \MakeLowercase{\textit{et al.}}: BeyondFusion}

\maketitle

\begin{abstract}

Mobile infrared-visible imaging typically pairs a compact infrared sensor with a high-resolution visible camera for complementary perception.
While cross-sensor misalignment caused by different optics, viewpoints, fields of view, and exposure timings hinders practical deployment.
In this paper, we propose BeyondFusion, a unified latent diffusion framework for calibration-free visible-guided infrared super-resolution and infrared-visible fusion tasks.
The proposed framework supports both task-specific training and joint training where two tasks are optimized and executed as two readouts of the same generative process.
Instead of relying on explicit registration or geometric warping, BeyondFusion introduces a cross-modal self-aligning (CMSA) module into the denoising U-Net.
CMSA reorganizes infrared and visible latent tokens into a shared attention space to learn content-adaptive cross-modal correspondence during the denoising process.
Together with misalignment augmentation module, the model is facilitated to exploit visible structural and semantic cues while preserving thermal consistency, enabling high-frequency infrared reconstruction and informative fused-image generation under uncalibrated conditions.
Extensive experiments on public benchmarks and a mobile infrared-visible imaging system show strong performance across aligned inputs, low-resolution infrared observations, synthetic misalignments, and real mobile captures with unsynchronized sensors.
Ablation studies, unified training analysis, and downstream pedestrian detection further validate the effectiveness of BeyondFusion for calibration-free multimodal imaging.
\end{abstract}


\input{sec/intro}

\input{sec/related_work}
\input{sec/methodology}
\input{sec/experiments}

\input{sec/conclusion}
\bibliographystyle{IEEEtran}
\bibliography{reference}



\end{document}

%% file: sec/intro.tex
\section{Introduction}
\IEEEPARstart{I}{nfrared} imaging captures thermal radiation emitted by objects, enabling perception beyond the visible spectrum and providing complementary cues for scene understanding.
It is particularly valuable in challenging environments such as low illumination, fog, smoke, and nighttime scenes, where conventional visible cameras are often unreliable.
Together with visible imaging, which provides rich color, texture, and semantic information, infrared sensing has become an important component of multimodal perception systems for safety-critical applications, including autonomous driving, robotic navigation, and situational awareness \cite{liu2025dcevo,ma2019infrared,tang2023divfusion,shin2023deep,tang2023happened,sheinin2024projecting}.

Recent mobile devices make infrared-visible imaging increasingly practical, but they also expose a fundamental hardware bottleneck.
Compared with mature visible cameras, compact infrared sensors are constrained by limited aperture size, the long wavelength of infrared radiation, manufacturing cost, and restricted pixel density.
As a result, mobile infrared cameras usually produce low-resolution images with blurry structures, weak boundaries, and limited textural detail.
This motivates infrared image super-resolution (SR), which aims to recover high-resolution thermal details from low-resolution infrared observations \cite{wang2020deep}.
However, single-image infrared SR is severely ill-posed because high-frequency details are lost during low-resolution infrared sensing.
This difficulty is further compounded by the scarcity of large-scale infrared datasets, which limits learning-based restoration and generalization.
Visible-guided infrared SR therefore becomes an appealing solution: high-resolution visible images can provide structural and semantic cues that are difficult to infer from infrared observations alone \cite{cnncsr,corefusion,swinfusr}.

Beyond enhancing infrared resolution, the same mobile infrared-visible sensing setup also calls for a representation that combines the complementary strengths of both modalities.
When the goal is human inspection or downstream perception rather than infrared restoration alone, a purely infrared output may still miss visible-domain appearance cues such as color, fine object contours, and non-thermal scene structures.
Infrared-visible image fusion (IVF), as a representative multimodal image fusion (MMIF) task, addresses this complementary need by preserving thermal saliency while incorporating visible structural and semantic information.
Driven by deep learning, IVF has progressed from CNN- and GAN-based methods to Transformer, hybrid CNN-Transformer, and diffusion-based frameworks \cite{IFCNN,U2Fusion,FusionGAN,SwinFusion,cddfuse,Mask-DiFuser,Dif-fusion,li2025towards}.
Together, visible-guided infrared SR and IVF represent two output forms of the same mobile dual-sensor imaging problem: one enhances the infrared modality itself, while the other produces a fused representation for perception-oriented use.
This observation suggests that the two tasks need not be formulated as isolated pipelines.

Despite their different outputs, these two tasks share the same central obstacle: reliable cross-modal interaction under real hardware conditions.
In a mobile imaging system, infrared and visible images are captured by separate sensors with different optics, fields of view, spatial resolutions, viewpoints, and exposure timings, making their spatial discrepancy device- and scene-dependent.
Meanwhile, the two modalities sense largely non-overlapping spectral ranges: infrared imaging captures thermal radiation, whereas visible imaging records reflected visible light.
Their observations therefore encode different physical properties, and corresponding regions may exhibit inconsistent structures, textures, or saliency across modalities.
As a result, effective cross-modal enhancement and fusion require more than pixel-level registration; the model must also adaptively determine which visible cues are helpful and how they should interact with thermal observations.
However, most existing visible-guided infrared restoration and IVF methods still follow a pixel-level fusion paradigm, assuming accurately aligned infrared-visible pairs or introducing explicit registration, geometric warping, deformation fields, or dedicated alignment modules \cite{MulFS-CAP,COPDR,wang2022unsupervised}.
Although effective in constrained settings, these strategies bind fusion quality to the accuracy of pixel-level correspondence estimation and become fragile when misalignment varies with device layout, scene depth, resolution, or temporal offset.
This raises a key question: can heterogeneous visible-infrared observations be organized within one imaging framework to support different low-level objectives without first forcing the two modalities into strict pixel correspondence?

Our answer is to move beyond task-specific network design and pixel-level fusion, and to formulate calibration-free visible-infrared imaging as interaction within a shared latent generative workspace.
Instead of enforcing one-to-one spatial correspondence, BeyondFusion represents infrared and visible inputs as latent tokens and lets their interaction emerge through shared self-attention during diffusion denoising.
This token-level formulation provides a soft, content-adaptive mechanism for cross-modal information exchange, making it less dependent on hardware calibration and naturally extensible to other multimodal sensing settings.

Based on this principle, we propose BeyondFusion, a unified self-aligned latent diffusion framework for calibration-free visible-guided infrared super-resolution (SR) and infrared-visible fusion (IVF).
Its core module, cross-modal self-aligning (CMSA), reorganizes infrared and visible latent tokens into a shared attention sequence, enabling inter-modal guidance while preserving modality-specific structures.
Together with misalignment-aware augmentation and one-step latent diffusion \cite{ADD}, BeyondFusion learns robust cross-modal interaction from imperfectly aligned data and produces high-fidelity infrared and fused images under real mobile conditions.
We validate the proposed framework on public benchmarks and a real mobile infrared-visible imaging system, where extensive experiments demonstrate strong performance in both infrared SR and IVF under aligned inputs, low-resolution infrared observations, uncalibrated acquisition, and downstream perception.

A preliminary conference version of this work appeared as 3M-TI \cite{3M}.
This manuscript substantially extends it by expanding the problem scope from calibration-free mobile infrared super-resolution to a unified framework for both visible-guided infrared SR and IVF, adapting the cross-modal diffusion framework to support super-resolution and fusion within a single framework, and providing new fusion experiments, fusion-specific benchmark evaluations, ablation analyses, and extended real mobile imaging validation for the fusion setting under uncalibrated conditions.

The main contributions of BeyondFusion can be summarized as follows:
\begin{itemize}
\item \textbf{Unified latent formulation for calibration-free multimodal imaging}: We formulate visible-guided infrared SR and infrared-visible fusion as two outputs of one latent generative framework, jointly addressing resolution degradation and cross-modal misalignment in mobile infrared-visible imaging.
\item \textbf{Attention-based latent workspace for implicit cross-modal interaction}: We propose CMSA to organize visible and infrared tokens in a shared self-attention space, enabling soft cross-modal interaction without explicit registration, geometric warping, or pixel-level correspondence estimation.
\item \textbf{Joint training and practical validation}: We show that infrared SR and infrared-visible fusion can be optimized and executed jointly with little performance sacrifice, and validate BeyondFusion on public benchmarks, simulated uncalibrated settings, a real mobile imaging system, and downstream pedestrian detection.
\end{itemize}

%% file: sec/related_work.tex
\section{Related Work}
We review the most relevant literature from three perspectives: reference-guided image restoration, infrared-visible image fusion, and diffusion-powered image enhancement.

\subsection{Reference-Guided Image Restoration}
Reference-guided image restoration leverages an auxiliary high-resolution image or modality to recover details that are missing from a degraded target observation.
This paradigm has been studied in visible image super-resolution \cite{zheng2018crossnet,tan2020crossnet++,zhang2019image,fang2023engram}, hyperspectral image super-resolution \cite{cnncsr,SSC-HSR}, and infrared image enhancement \cite{swinfusr,swinpaste}.
For infrared restoration, visible-guided methods are particularly attractive because visible images provide structural, textural, and semantic cues that are difficult to infer from low-resolution infrared inputs alone.
Early approaches such as CoReFusion \cite{corefusion} adopt dual-encoder architectures to extract and fuse cross-modal features, with contrastive constraints introduced to improve modality consistency.
Transformer-based methods further strengthen long-range dependency modeling and cross-modal interaction through attention mechanisms \cite{attention}.
Representative methods include MGNet \cite{MGNet}, which mines multi-level visible cues for UAV infrared super-resolution, SwinFuSR \cite{swinfusr}, which introduces a lightweight Swin Transformer backbone, SwinPaste \cite{swinpaste}, which improves training via data mixing and multi-scale supervision, and MSFFCT \cite{MSFFCT}, which combines channel-wise Transformers with multi-scale feature aggregation.

Despite these advances, most visible-guided infrared restoration methods are still developed for aligned visible-infrared pairs \cite{corefusion,MGNet,swinfusr,swinpaste,MSFFCT}.
When visible and infrared images are captured by diverse uncalibrated mobile sensors, device-dependent viewpoint, resolution, and timing discrepancies can lead to substantial and spatially varying misalignment.
This motivates an automatic restoration framework that directly accepts unaligned infrared-visible inputs and exploits visible guidance without a separate calibration, registration, or warping stage.

\subsection{Infrared-Visible Image Fusion}
Infrared-visible image fusion aims to generate an informative representation that preserves thermal saliency from infrared images while incorporating structural and semantic details from visible images.
Existing methods can be roughly grouped into task-driven, semantic-driven, and general-purpose fusion frameworks.
Task-driven methods \cite{risfuse,cdtfusion,freefusion} optimize fused representations for downstream perception, while semantic-driven methods \cite{liu2026fine,MDDPFuse,SPDFusion} introduce object-level or scene-level guidance to improve high-level consistency.
General fusion models, such as CDDFuse \cite{cddfuse} and SwinFusion \cite{SwinFusion}, combine CNNs, Transformers, or attention mechanisms to model complementary information across modalities.

However, many fusion methods are designed for well-registered infrared-visible pairs, and unregistered inputs can lead to duplicated edges, structural distortions, or weakened thermal-visible correspondence.
Recent methods address this issue by explicitly modeling cross-modal correspondence within the fusion process.
MulFS-CAP \cite{MulFS-CAP} constructs an alignment perception matrix to reorganize local features, while C-OPDR \cite{COPDR} estimates deformation fields before fusion.
In contrast, our goal is to let cross-modal interaction emerge automatically through latent self-attention in diffusion, without estimating an alignment matrix, deformation field, or geometric warp.

\subsection{Diffusion-Powered Image Enhancement}
Diffusion models have shown strong generative priors for image restoration, super-resolution, and image enhancement, especially in recovering realistic structures and high-frequency details \cite{diffusion,ma2026taming,dong2026dreamsr}.
Latent diffusion methods perform denoising in the compact latent space of a variational autoencoder and have been widely adopted in perceptual restoration and super-resolution \cite{dmdiff,seesr,sinsr,osediff}.
One-step diffusion models further simplify the sampling process by generating restored images with a single denoising step \cite{ADD}.
These advances have also been introduced into infrared enhancement and image fusion.
For example, DifIISR \cite{difiisr} incorporates gradient-based priors and infrared spectrum distribution regularization to model infrared-specific frequency characteristics, while Mask-DiFuser \cite{Mask-DiFuser} reformulates unsupervised fusion as reconstruction from dual-masked inputs.
More recently, DS2D~\cite{wu2026ds2d} introduces a state-space diffusion framework for infrared-visible image fusion, decoupling feature guidance to better preserve modality-specific information and enhance cross-modal complementary representation.
ReCoFuse~\cite{ReCoFuse} further explores restorative multi-modal diffusion with reciprocal coupling, where modality-specific restoration branches and time-aware cross-modal integration modules are jointly optimized to improve fusion robustness under complex degradations.

The conference version of this work, 3M-TI \cite{3M}, explored calibration-free visible guidance for mobile infrared super-resolution within this diffusion-powered enhancement line.
This extension broadens that setting from infrared super-resolution alone to a unified framework for both visible-guided infrared super-resolution and infrared-visible fusion.
BeyondFusion focuses on how heterogeneous visible-infrared observations can be organized within a shared latent workspace and decoded for different imaging objectives.

%% file: sec/methodology.tex
\section{Methodology}
BeyondFusion instantiates a unified latent generative formulation for calibration-free visible-infrared imaging.
Given a visible-infrared pair captured without geometric calibration or hardware synchronization, it organizes the two observations into a shared latent workspace in which visible structures and infrared thermal responses can interact.
The resulting latent representation is decoded into infrared super-resolution and infrared-visible fusion outputs within one framework.

\begin{figure*}[htbp]
    \centering
    \includegraphics[width=0.9\textwidth]{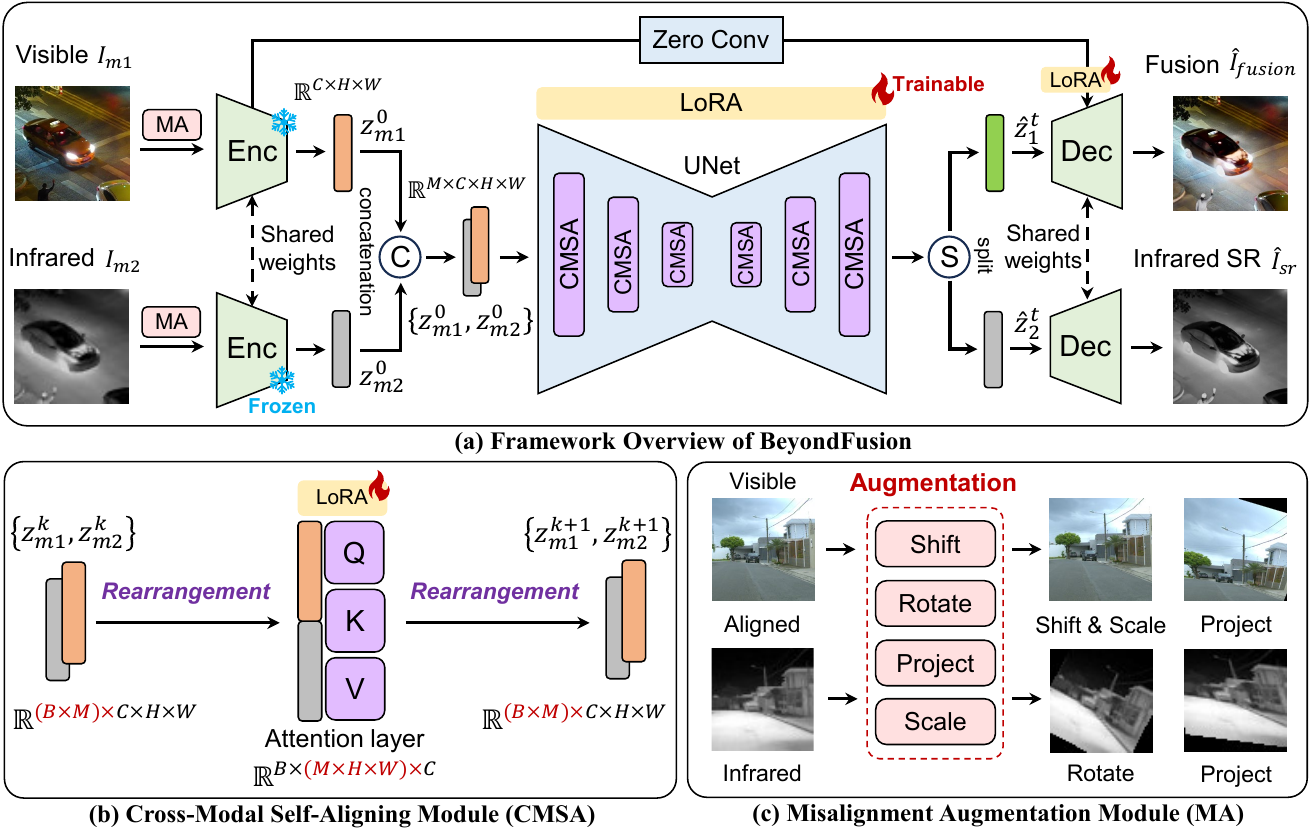}
    \caption{Overview of the BeyondFusion architecture. BeyondFusion encodes visible and infrared inputs into a latent diffusion space, reshapes their latent representations through the cross-modal self-aligning module (CMSA), and decodes the refined latents for either SR or IVF. LoRA fine-tuning is applied to both the UNet and the VAE decoder. (b) CMSA. Two reshape operations are applied before and after the original self-attention layers, allowing visible and infrared tokens to interact in a shared attention sequence without predicting explicit alignment fields. (c) Misalignment Augmentation Module (MA). Controlled geometric perturbations are applied to improve robustness against camera parallax and temporal mismatch.}
    \label{fig:method_framework}
\end{figure*}

\subsection{BeyondFusion: A Unified Latent-Space Formulation}
\label{overview}
BeyondFusion treats the pretrained diffusion model as a shared latent workspace for heterogeneous observations, rather than as a task-specific restoration or fusion network.
Given an infrared image \(I_{\mathrm{ir}}\) and a visible image \(I_{\mathrm{vis}}\), the VAE encoder \(E\) first maps them into latent features and directly folds the modality dimension into the batch dimension:
\begin{equation}
    z^0=\mathrm{fold}\big(\mathrm{stack}(E(I_{\mathrm{vis}}), E(I_{\mathrm{ir}}))\big)
    \in \mathbb{R}^{(B M) \times C \times H \times W},
\end{equation}
where \(B\) is the batch size, \(M=2\) denotes the visible and infrared modalities, and \(C\), \(H\), and \(W\) are the channel and spatial dimensions.
This folded representation has the standard input shape expected by the pretrained diffusion U-Net, so the original network layers can be reused without changing their dimensions.

The folded latent \(z^0\) is then processed by the one-step denoising U-Net adapted from SD-Turbo~\cite{ADD}.
In most layers, the visible and infrared latents are processed as independent samples in the folded batch, without cross-modal information exchange.
At the self-attention layers, CMSA reshapes the latent tokens from the two modalities into a shared attention sequence, allowing the visible and infrared latent images to exchange information.
This design turns the original denoising process into a cross-modal latent workspace without a separate registration network and without redesigning the U-Net backbone.
In practice, low-rank adaptation (LoRA)~\cite{lora} is applied to the U-Net and VAE decoder for efficient adaptation.

After denoising, the VAE decoder maps the refined latent representation to the task output.
This provides a high-resolution infrared image for SR and a fused RGB image for IVF, treating the two tasks as different readouts of the same latent cross-modal generative process, as shown in Fig.~\ref{fig:method_framework}(a).

\subsection{Cross-Modal Self-Aligning Module}
\label{sec:csa}
CMSA introduces cross-modal interaction into the existing self-attention (SA) layers of the pretrained U-Net, as shown in Fig.~\ref{fig:method_framework}(b).
For the \(l\)-th SA layer, let \(z_l \in \mathbb{R}^{(B M) \times C_l \times H_l \times W_l}\) denote the folded latent feature.
CMSA restores the modality dimension and flattens visible and infrared spatial tokens into one joint sequence:
\begin{equation}
\begin{aligned}
    z_l^{\mathrm{seq}} &= \mathrm{reshape}(z_l)
    \in \mathbb{R}^{B \times (M H_l W_l) \times C_l}, \\
    \bar{z}_l^{\mathrm{seq}} &= \mathrm{SA}(z_l^{\mathrm{seq}}), \\
    \bar{z}_l &= \mathrm{reshape}(\bar{z}_l^{\mathrm{seq}})
    \in \mathbb{R}^{(B M) \times C_l \times H_l \times W_l}.
\end{aligned}
\end{equation}
Here, \(\mathrm{SA}\) is the original self-attention operation: the query, key, value, and output projections are the same projections already present in the U-Net, with LoRA updates used during fine-tuning.
After SA, the output is reshaped back to the folded latent layout, so the following U-Net layers can operate as usual.
Thus, CMSA turns existing SA layers into cross-modal workspace interaction layers without adding learnable projections, attention heads, or alignment modules.

The main advantage is that cross-modal correspondence is represented implicitly through attention, rather than as an explicit geometric transform.
Alignment-based methods often predict an alignment matrix, deformation field, or warp~\cite{MulFS-CAP,COPDR,wang2022unsupervised}; inaccurate estimates can propagate into the output as ghosting, duplicated edges, or distorted structures.
CMSA does not impose any predicted alignment on the two modalities; instead, visible and infrared tokens interact through soft attention inside the diffusion denoising process.
This lets the pretrained generative backbone integrate complementary cross-modal cues within the shared latent workspace while reducing artifacts caused by inaccurate explicit alignment.

\subsection{Decoding for Super-Resolution and Image Fusion}
\label{sec:decoding}
The denoising U-Net outputs a folded latent tensor \(z^\star \in \mathbb{R}^{(B M) \times C \times H \times W}\), which has the same layout as the input latent \(z^0\).
Before decoding, this tensor is split into two output streams under task-specific supervision:
\begin{equation}
\begin{aligned}
    (z^\star_{\mathrm{sr}}, z^\star_{\mathrm{fusion}})
    &= \mathrm{unfold}(z^\star).
\end{aligned}
\end{equation}
Although the input folding is organized by modality, the denoised latent entries are decoded as task-specific streams, where \(z^\star_{\mathrm{sr}}\) is used for infrared super-resolution and \(z^\star_{\mathrm{fusion}}\) is used for image fusion.
The two streams share the same VAE decoder \(D\), with output-specific skip features:
\begin{equation}
\begin{aligned}
    \hat{I}_{\mathrm{sr}}
    &= D(z^\star_{\mathrm{sr}}; f_{\mathrm{ir}}),\\
    \hat{I}_{\mathrm{fusion}}
    &= D(z^\star_{\mathrm{fusion}}; f_{\mathrm{vis}}),
\end{aligned}
\end{equation}
where \(f_{\mathrm{ir}}\) denotes infrared feature maps extracted from the infrared input for super-resolution, and \(f_{\mathrm{vis}}\) denotes visible feature maps used to preserve structural and color fidelity in the fusion output.
The infrared response required for image fusion is incorporated through the cross-modal latent stream \(z^\star_{\mathrm{fusion}}\).
Following~\cite{parmar2024one}, feature maps from four encoder downsampling blocks are passed through zero-initialized \(1\times1\) convolutions and added to the corresponding decoder upsampling blocks.
This design enables BeyondFusion to obtain infrared super-resolution and image fusion outputs from a shared cross-modal diffusion process and a shared VAE decoder, without introducing task-specific decoding branches.

\subsection{Misalignment-Aware Training}
\label{sec:augmentation}
Infrared-visible image pairs are acquired by physically distinct sensors, so their raw observations are generally affected by camera parallax, different fields of view, resolution mismatch, and exposure-time differences.
Public datasets often reduce these discrepancies through careful calibration, rectification, cropping, or registration, making the data convenient for training and evaluation but also creating a gap between benchmark pairs and raw sensor observations.
Methods trained under such assumptions can therefore become difficult to deploy across diverse visible-infrared imaging systems.
To expose BeyondFusion to such conditions, we apply controlled geometric perturbations to the input modalities during training, including translation, scaling, rotation, and perspective warping, as shown in Fig.~\ref{fig:method_framework}(c).
These perturbations are not intended to reproduce the exact distribution of real sensor misalignment.
Rather, since CMSA models cross-modal correspondence implicitly through latent attention and does not require pixel-level registration, synthetic perturbations are sufficient to discourage overfitting to fixed pixel neighbors and to encourage robust cross-modal interaction under spatial discrepancies.

\subsection{Mobile Infrared-Visible Imaging System}
\label{sec:imaging_system}
\begin{figure}[ht]
    \centering 
    \includegraphics[width=0.85\columnwidth]{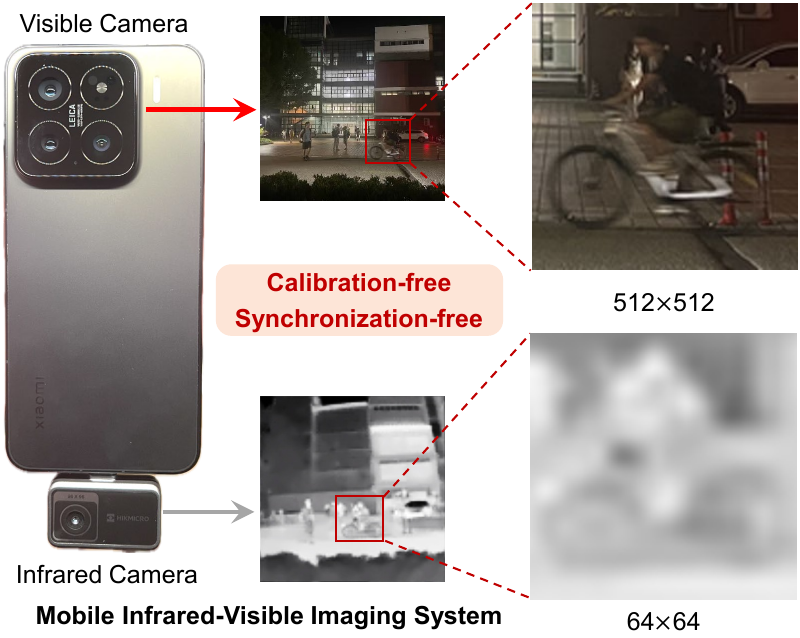}
    \caption{Mobile infrared-visible imaging system.}
    \label{fig:imaging_system}
\end{figure}

To validate BeyondFusion under practical mobile imaging conditions, we build a portable infrared-visible acquisition system by attaching a compact infrared camera to a commercial smartphone, as shown in Fig.~\ref{fig:imaging_system}.
The visible and infrared images are captured by two independent sensors without geometric calibration or hardware synchronization.
As a result, the paired observations naturally differ in viewpoint, field of view, spatial resolution, and capture timing.

The infrared module provides a native resolution of \(96 \times 96\), which is resized to \(64 \times 64\) in this work for a standardized experimental setting, with a \(\ang{50} \times \ang{50}\) field of view (FOV) and a pixel pitch of \SI{12}{\micro\meter}.
For visible image collection, we use the primary camera of a Xiaomi 15 smartphone, which is equipped with an OV50H sensor and captures images at a resolution of \(4096 \times 4096\).
This visible camera has an equivalent focal length of 23~mm and a FOV of \(\ang{74} \times \ang{59}\), which is wider than that of the infrared module.

%% file: sec/experiments.tex
\section{Experiments}

We design the experiments to validate BeyondFusion as a unified framework for calibration-free visible-infrared imaging.
Specifically, we examine whether the same latent generative formulation can support infrared super-resolution and infrared-visible image fusion, whether the latent interaction remains robust to resolution gaps and uncalibrated inputs, and whether the fused outputs benefit downstream perception.
\subsection{Experimental Setup}
\subsubsection{Datasets and Preprocessing}
For infrared super-resolution, we use IRVI~\cite{IRVI}, LLVIP~\cite{LLVIP}, M$^3$FD~\cite{M3FD}, and the PBVS 2025 TISR Challenge Track 2~\cite{PBVS25}, yielding 10,922 training pairs and 1,176 test pairs.
Official splits are followed when available, and M$^3$FD is partitioned with a disjoint sampling protocol.
For infrared-visible image fusion, we use MSRS~\cite{MSRS}, M$^3$FD, LLVIP, and M3SVD~\cite{VideoFusion}, yielding 10,366 training pairs and 1,409 test pairs.

Since the SD-Turbo backbone and VAE are operated with \(512 \times 512\) three-channel inputs, all visible and infrared images are first center-cropped to a square region and mapped to this canonical input size before latent encoding.
For super-resolution, the visible image is resized to \(512 \times 512\) and used as the guidance input.
The low-resolution infrared condition is generated by downsampling the infrared image to \(64 \times 64\), adding Gaussian noise to approximate compact infrared sensors, and resizing it back to \(512 \times 512\) for network input; the corresponding high-resolution infrared image is used as the reconstruction target.
For image fusion, infrared inputs are prepared at \(512 \times 512\), \(128 \times 128\), and \(64 \times 64\) and then resized to \(512 \times 512\), allowing us to evaluate fusion robustness under different source infrared resolutions while keeping the diffusion input size fixed.
Single-channel infrared images are duplicated into three channels to match the VAE input format.
The misalignment perturbations described in Section~\ref{sec:augmentation} are applied during training and are also used to construct synthetic uncalibrated test settings when evaluating robustness to spatial offsets and scale variations.

\subsubsection{Mobile Dataset}
We further collect 200 image pairs from 80 scenes using the mobile infrared-visible imaging system described in Section~\ref{sec:imaging_system}.
Most scenes are captured outdoors at night and are used to evaluate real uncalibrated mobile acquisition.
For these captures, visible images are center-cropped and resized to \(512 \times 512\), while native \(96 \times 96\) infrared images are directly resized to \(64 \times 64\).

\subsubsection{Implementation Details}
BeyondFusion can be trained as a single unified model that predicts \(\hat{I}_{\mathrm{sr}}\) and \(\hat{I}_{\mathrm{fusion}}\) in one forward pass.

For infrared super-resolution, the high-resolution infrared image \(I_{\mathrm{ir}}^{\mathrm{hr}}\) is used as the reconstruction target, and the loss function is defined as
\begin{align}
    \mathcal{L}_{\mathrm{sr}} = \lambda_{L_2}^{sr}\mathcal{L}_{2} + \mathcal{L}_{\mathrm{LPIPS}}.
    \label{Loss_sr}
\end{align}
Here, \(\mathcal{L}_{2}\) and \(\mathcal{L}_{\mathrm{LPIPS}}\) are computed between \(\hat{I}_{\mathrm{sr}}\) and \(I_{\mathrm{ir}}^{\mathrm{hr}}\).

For image fusion, let \(I_{\mathrm{fusion}}\) denote the fused supervision generated by Mask-Difuser~\cite{Mask-DiFuser}, \(\hat{I}_{\mathrm{fusion}}\) denote the predicted fused image, and \(I_{\mathrm{vis}}\) denote the input visible image.
The loss function is defined as
\begin{equation}
\begin{aligned}
    \mathcal{L}_{\mathrm{fusion}} &=
    \mathcal{L}_{\mathrm{int}}
    + \mathcal{L}_{\mathrm{color}}
    + \lambda_{L_2}^{fu}\mathcal{L}_{2}
    + \mathcal{L}_{\mathrm{LPIPS}}, \\
    \mathcal{L}_{\mathrm{int}} &=
    \left\|Y(\hat{I}_{\mathrm{fusion}}) - Y(I_{\mathrm{fusion}})\right\|_1, \\
    \mathcal{L}_{\mathrm{color}} &=
    \left\|C(\hat{I}_{\mathrm{fusion}}) - C(I_{\mathrm{vis}})\right\|_1,
\end{aligned}
\label{Loss_fusion}
\end{equation}
where \(\mathcal{L}_{int}\), \(\mathcal{L}_{2}\), and \(\mathcal{L}_{\mathrm{LPIPS}}\) are computed between \(\hat{I}_{\mathrm{fusion}}\) and \(I_{\mathrm{fusion}}\), \(Y(\cdot)\) extracts the intensity channel of YUV space;
\(\mathcal{L}_{color}\) is calculated between \(\hat{I}_{\mathrm{fusion}}\) and \(I_{\mathrm{vis}}\), \(C(\cdot)\) extracts the color channels.
We set \(\lambda_{L_2}^{sr}=10\), \(\lambda_{L_2}^{fu}=2\), and the remaining weights each to 1.
The total objective for joint training is
\begin{equation}
    \mathcal{L}_{\mathrm{total}} =
    \mathcal{L}_{\mathrm{sr}} + \mathcal{L}_{\mathrm{fusion}}.
    \label{Loss_total}
\end{equation}

We further use classifier-free guidance (\textit{CFG})~\cite{cfg} in the unified denoising process.
Following the common \textit{CFG} practice, BeyondFusion trains positive and negative denoising branches and combines their predictions at inference.
For the positive branch, the model is supervised by the aforementioned targets.
For the negative branch, $\mathcal{L}_{\mathrm{int}}$ and $\mathcal{L}_{\mathrm{color}}$ in Eq.~(\ref{Loss_fusion}) are removed.
Besides, we replace \(I_{\mathrm{ir}}^{\mathrm{hr}}\) with the low-resolution infrared input \(I_{\mathrm{ir}}^{\mathrm{lr}}\) in Eq.~(\ref{Loss_sr}) and replace \(I_{\mathrm{fusion}}\) with the visible image \(I_{\mathrm{vis}}\) in Eq.~(\ref{Loss_fusion}).
These two replacements correspond to non-super-resolved and non-fused targets, respectively.
Positive and negative samples are drawn at a ratio of 7:3.
During inference, the guided prediction is
\begin{align}
    \tilde{\boldsymbol{\epsilon}} = \boldsymbol{\epsilon}_{neg} + \omega \cdot \left(\boldsymbol{\epsilon}_{pos} - \boldsymbol{\epsilon}_{neg} \right),
    \label{cfg}
\end{align}
where $\boldsymbol{\epsilon}_{pos}$ and $\boldsymbol{\epsilon}_{neg}$ are the positive and negative predictions of the unified model, and $\omega$ is the guidance scale.
The combination in Eq.~(\ref{cfg}) is performed in the latent image space before VAE decoding, and the guided latent is decoded into \(\hat{I}_{\mathrm{sr}}\) and \(\hat{I}_{\mathrm{fusion}}\) jointly.
We apply $\omega=1.1$ in all inferences.

All our models are trained with the Adam optimizer~\cite{Adam}, using a learning rate of $2\times10^{-5}$ and a batch size of 4 on a single NVIDIA A800 (80~GB) GPU.
LoRA is applied with rank 16 for the U-Net and rank 4 for the VAE decoder.

\subsubsection{Comparative Methods}
For infrared super-resolution, we compare BeyondFusion with CoReFusion~\cite{corefusion}, CoRPLE~\cite{corple}, SwinFuSR~\cite{swinfusr}, SwinPaste~\cite{swinpaste}, SeeSR~\cite{seesr}, OSEDiff~\cite{osediff}, and DifIISR~\cite{difiisr}, covering visible-guided, infrared-specific, and diffusion-based restoration baselines.
For infrared-visible image fusion, we compare with CDDFuse~\cite{cddfuse}, Mask-Difuser~\cite{Mask-DiFuser}, and C-OPDR~\cite{COPDR}.
All baselines are retrained on our datasets using their released implementations.

\subsubsection{Evaluation Metrics}
For infrared super-resolution, we report PSNR and SSIM~\cite{ssim} for reconstruction fidelity, LPIPS~\cite{lpips} for perceptual similarity, and MUSIQ and MANIQA~\cite{IQA-pytorch} for no-reference quality assessment.
For infrared-visible image fusion, we report entropy (EN), standard deviation (SD), spatial frequency (SF), structural content difference (SCD), and average gradient (AG), following the common definitions in~\cite{ma2019infrared}.

Unless otherwise specified, BeyondFusion is compared with baselines using a task-specific training pipeline, where Eq.~(\ref{Loss_sr}) supervises infrared super-resolution and Eq.~(\ref{Loss_fusion}) trains infrared-visible image fusion.
Each model is with the \textit{CFG} strategy introduced.
The impact of the training pipeline is further analyzed in Section~\ref{train_pipeline}.

\begin{figure*}[t]
    \centering 
    \includegraphics[width=\textwidth]{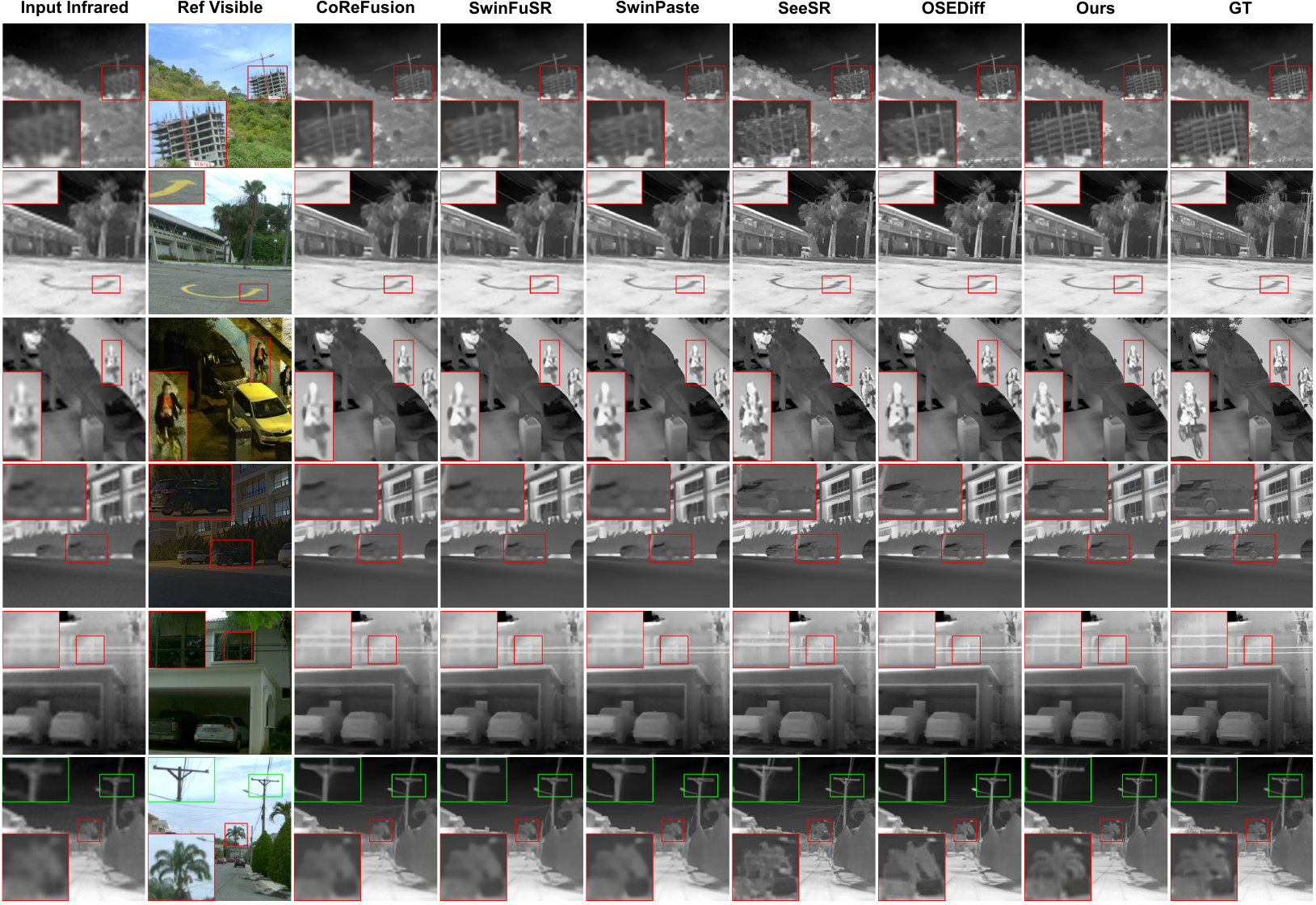}
    \caption{Qualitative comparison for infrared image super-resolution on the test set. BeyondFusion reconstructs sharper structures and more faithful infrared patterns, with details that are visually plausible and better aligned with the ground truth.}
    \label{fig:sota_sr}
\end{figure*}

\begin{figure*}[t]
    \centering 
    \includegraphics[width=\textwidth]{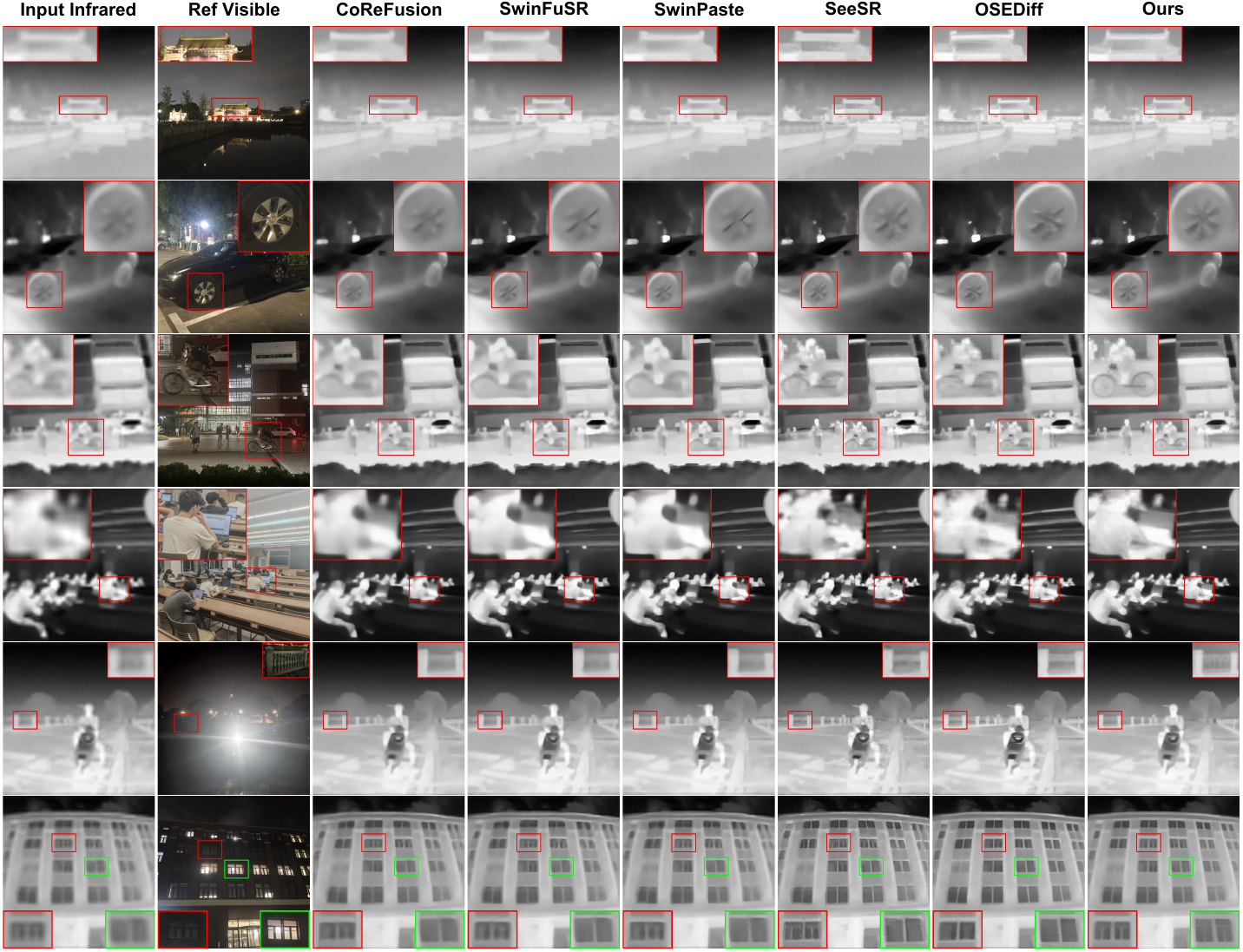}
    \caption{Qualitative infrared super-resolution comparison on the real-world mobile dataset. BeyondFusion shows strong generalization to uncalibrated captures, producing sharper and more faithful infrared details from imperfect visible references.}
    \label{fig:real_val_sr}
\end{figure*}

\subsection{Infrared Image Super-Resolution}
\label{sec:exp_sr}
Table~\ref{tab:sota_sr} compares BeyondFusion with visible-guided, infrared-specific, and diffusion-based super-resolution methods on public infrared-visible benchmarks.
BeyondFusion achieves the best LPIPS score and the second-best rest scores and keeps PSNR and SSIM close to the strongest fidelity-oriented baseline, CoRPLE.
Compared with visible-guided baselines such as SwinFuSR and SwinPaste, BeyondFusion reduces LPIPS by about 0.14 and improves MANIQA by about 0.18, showing a clear perceptual advantage.
Compared with diffusion restorers OSEDiff and DifIISR, BeyondFusion gives substantially higher fidelity, improving PSNR by 2.33 dB and 2.10 dB, respectively.
These results indicate BeyondFusion's superiority in balancing perceptual quality and structural fidelity.

The effect of misalignment-aware augmentation is also reflected in Table~\ref{tab:sota_sr}.
With the same augmentation protocol, BeyondFusion effectively improves the results in terms of all metrics.
By contrast, the same augmentation brings fewer improvements to other visible-guided baselines.

\begin{table}[htbp]
  \centering
  \caption{Quantitative comparison on infrared image super-resolution. Gray cells indicate the best result, and light gray cells indicate the second-best result for each metric.}
  \footnotesize
  \setlength{\tabcolsep}{1pt}
    \begin{tabular}{c|ccccc}
    \hline
    \diagbox[width=10em, height=1.8em, innerleftsep=0pt, innerrightsep=0pt]{Methods}{Metrics} & PSNR$\uparrow$ & SSIM$\uparrow$ & LPIPS$\downarrow$ & MANIQA$\uparrow$ & MUSIQ$\uparrow$ \\
    \hline
    CoReFusion \cite{corefusion} & 30.11  & 0.8588  & 0.3214  & 0.2771  & 28.35  \\
    CoReFusion w/ Augment & 30.30 & 0.8634 & 0.3174  & 0.2748  & 28.95  \\
    \hline
    CoRPLE \cite{corple} & \cellcolor[rgb]{ .502,  .502,  .502}\textbf{30.47} & \cellcolor[rgb]{ .502,  .502,  .502}\textbf{0.8642} & 0.3206 & 0.2833 & 30.46 \\
    \hline
    SwinFuSR \cite{swinfusr} & 29.85  & 0.8549  & 0.3085  & 0.2740  & 29.86  \\
    SwinFuSR w/ Augment & 29.98  & 0.8581  & 0.3134  & 0.2753  & 30.33  \\
    \hline
    SwinPaste \cite{swinpaste} & 29.83  & 0.8545  & 0.3075  & 0.2719  & 29.63  \\
    SwinPaste w/ Augment & 29.91  & 0.8575  & 0.3084  & 0.2745  & 30.66  \\
    \hline
    SeeSR \cite{seesr} & 29.41  & 0.8495  & 0.1828  & 0.4278  & 35.22  \\
    \hline
    OSEDiff \cite{osediff} & 28.05  & 0.8422  & 0.2113  & 0.4014  & 36.30 \\
    \hline
    DifIISR \cite{difiisr} & 28.28 & 0.8207 & 0.2754 & \cellcolor[rgb]{ .502,  .502,  .502}\textbf{0.5629} & \cellcolor[rgb]{ .502,  .502,  .502}\textbf{47.87} \\
    \hline
    BeyondFusion w/o Augment & 30.31 & 0.8632  & \cellcolor[rgb]{ .749,  .749,  .749}0.1820  & 0.4154  & 35.40 \\
    BeyondFusion w/ Augment & \cellcolor[rgb]{ .749,  .749,  .749}30.38  & \cellcolor[rgb]{ .749,  .749,  .749}0.8638  & \cellcolor[rgb]{ .502,  .502,  .502}\textbf{0.1658} & \cellcolor[rgb]{ .749,  .749,  .749}0.4536 & \cellcolor[rgb]{ .749,  .749,  .749}37.48 \\
    \hline
    \end{tabular}%
  \label{tab:sota_sr}%
\end{table}%

The qualitative comparison in Fig.~\ref{fig:sota_sr} further illustrates this behavior.
Non-diffusion methods often produce smooth infrared reconstructions with weakened high-frequency structures, while diffusion-based methods may synthesize sharp but structurally inconsistent details.
In Row 1, the steel frame of the building is blurred by CoReFusion, SwinFuSR, and SwinPaste, while SeeSR and OSEDiff recover sharper but inconsistent local patterns; BeyondFusion reconstructs clearer floor lines and vertical supports that are closer to the ground truth.
In Row 2, the road arrow remains weak or distorted in most baselines, whereas BeyondFusion restores a more recognizable arrow shape.
The utility pole and palm tree in the last row show a similar trend: BeyondFusion preserves object contours and thin structures with fewer hallucinated details than SeeSR and OSEDiff.
These examples support the quantitative observation that the method improves perceptual realism without drifting away from the target content.

We further evaluate generalization on the uncalibrated real mobile dataset in Fig.~\ref{fig:real_val_sr} and Table~\ref{tab:val}.
Similarly, conventional visible-guided baselines tend to produce blurry outputs, and diffusion restorers introduce visible artifacts in some regions.
BeyondFusion reconstructs sharper and more coherent infrared structures when the visible reference is spatially misaligned or degraded by light flare.
For instance, the wheel spokes in Row 2 and the bicycle structure in Row 3 are more distinguishable in BeyondFusion, while several baselines either blur the circular shape or introduce fragmented high-frequency artifacts.
In the last row, BeyondFusion also recovers clearer window grids from a strongly degraded nighttime capture.
The no-reference scores in Table~\ref{tab:val} are consistent with these observations, where BeyondFusion obtains the highest MUSIQ and MANIQA values on the mobile dataset.

\begin{table}[htbp]
  \centering
  \caption{No-reference image quality comparison on the real-world mobile infrared dataset. Gray cells indicate the best results.}
  \scriptsize
  \setlength{\tabcolsep}{1.5pt}
    \begin{tabular}{c|cccccc}
    \hline
    \diagbox{Metrics}{Methods} & CoReFusion & SwinFuSR & SwinPaste & SeeSR & OSEDiff & BeyondFusion \\
    \hline
    MUSIQ$\uparrow$ & 25.74  & 25.99  & 26.17 & 30.15 & 29.85  & \cellcolor[rgb]{ .502,  .502,  .502}\textbf{30.84} \\
    MANIQA$\uparrow$ & 0.2701  & 0.2754  & 0.2749 & 0.3454 & 0.3285  & \cellcolor[rgb]{ .502,  .502,  .502}\textbf{0.3562} \\
    \hline
    \end{tabular}%
  \label{tab:val}%
\end{table}%

\subsection{Infrared-Visible Image Fusion}
\label{sec:exp_fusion}
We evaluate image fusion in three settings that separate alignment, resolution gap, and data realism.
1) Simulated aligned setting: registered inputs with controllable infrared-visible resolution gaps, compared with fusion baselines that assume pixel-level alignment (CDDFuse~\cite{cddfuse} and Mask-Difuser~\cite{Mask-DiFuser}).
2) Simulated misaligned setting: unregistered inputs with controllable infrared-visible resolution gaps, compared with fusion baselines designed for unregistered inputs (C-OPDR~\cite{COPDR}).
3) Real mobile setting: used to assess practical generalization.

\begin{figure}[ht]
    \centering 
    \includegraphics[width=\columnwidth]{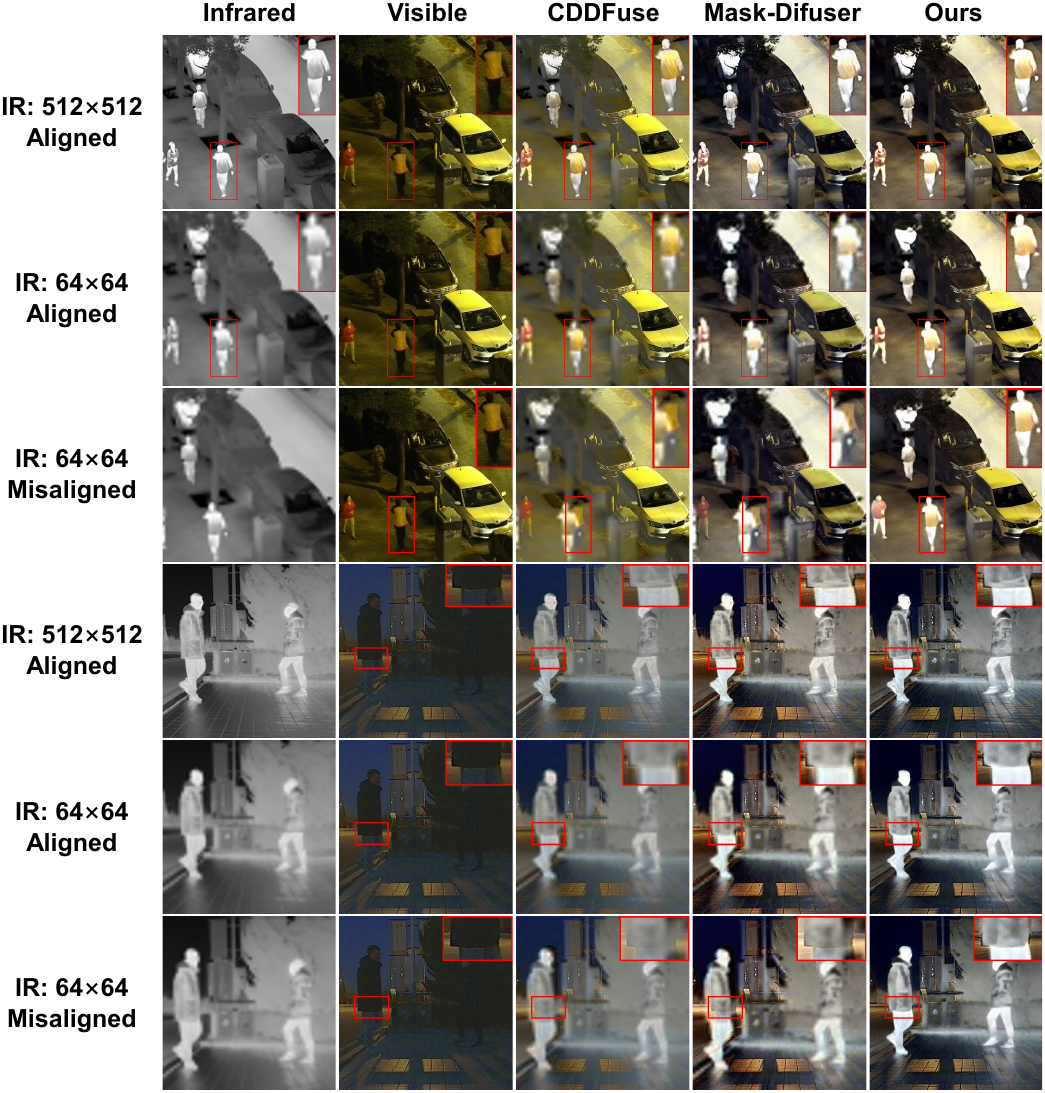}
    \caption{Qualitative comparison on our test set (zoom in for details). BeyondFusion maintains sharp edges and clear texture details even with downsampled infrared inputs.}
    \label{fig:sota_fusion_align}
\end{figure}

We first evaluate the simulated aligned setting, with quantitative results reported in Table~\ref{tab:sota_fu_align}.
When both modalities are aligned and available at \(512\times512\), BeyondFusion performs comparably to the aligned fusion baselines.
Its advantage becomes clearer as the infrared input is downsampled: at \(128\times128\) and \(64\times64\), BeyondFusion achieves the best scores on all reported metrics except SCD.
This trend suggests that BeyondFusion is less sensitive to enlarged infrared-visible resolution gaps, benefiting from visible structural guidance and the latent diffusion backbone to recover details absent from low-resolution infrared observations.
The qualitative comparison in Fig.~\ref{fig:sota_fusion_align} provides a more direct assessment.
Rows 1--3 and Rows 4--6 show two representative scenes under full-resolution aligned, low-resolution aligned, and low-resolution misaligned infrared inputs.
As shown in the enlarged person regions in Rows 1--3, the fused human regions produced by BeyondFusion remain visually consistent when the infrared input changes from \(512\times512\) to \(64\times64\).
In contrast, CDDFuse and Mask-Difuser lose fine infrared details under \(64\times64\) infrared inputs, leading to noticeably blurred human regions.
The degradation becomes more pronounced when low infrared resolution is coupled with misalignment: CDDFuse and Mask-Difuser produce shifted or ghosted responses around the human region, whereas BeyondFusion still maintains coherent fusion results.
Rows 4--6 further support this observation, where BeyondFusion better preserves human boundaries and clothing textures under both resolution degradation and spatial misalignment.
These observations indicate that BeyondFusion is more robust to degraded infrared details and cross-modal misalignment.

\begin{table}[htbp]
  \centering
  \caption{Performance comparison of different methods on aligned infrared-RGB image fusion. Gray cells indicate the best result.}
  \setlength{\tabcolsep}{1.7pt}
    \begin{tabular}{c|c|ccccc}
    \hline
    Methods & \diagbox{Resolution}{Metrics} & EN$\uparrow$  & SD$\uparrow$  & SF$\uparrow$  & SCD$\uparrow$ & AG$\uparrow$ \\
    \hline
    CDDFuse & \multirow{3}{*}{512$\times$512} & 7.067  & 49.480  & 16.892  & 1.661  & 4.792  \\
    Mask-Difuser &       & 7.284  & 60.048  & \cellcolor[rgb]{ .651,  .651,  .651}\textbf{18.715 } & \cellcolor[rgb]{ .651,  .651,  .651}\textbf{1.848 } & \cellcolor[rgb]{ .651,  .651,  .651}\textbf{5.749 } \\
    Ours  &       & \cellcolor[rgb]{ .651,  .651,  .651}\textbf{7.358} & \cellcolor[rgb]{ .651,  .651,  .651}\textbf{61.776} & 17.464  & 1.782  & 5.464  \\
    \hline
    CDDFuse & \multirow{3}{*}{128$\times$128} & 7.065  & 49.177  & 15.763  & 1.617  & 4.425  \\
    Mask-Difuser &       & 7.288  & 60.591  & 16.514  & \cellcolor[rgb]{ .651,  .651,  .651}\textbf{1.811 } & 5.124  \\
    Ours  &       & \cellcolor[rgb]{ .651,  .651,  .651}\textbf{7.367} & \cellcolor[rgb]{ .651,  .651,  .651}\textbf{61.600} & \cellcolor[rgb]{ .651,  .651,  .651}\textbf{16.848} & 1.766  & \cellcolor[rgb]{ .651,  .651,  .651}\textbf{5.266} \\
    \hline
    CDDFuse & \multirow{3}{*}{64$\times$64} & 7.058  & 48.688  & 15.422  & 1.561  & 4.186  \\
    Mask-Difuser &       & 7.291  & 60.927  & 15.714  & \cellcolor[rgb]{ .651,  .651,  .651}\textbf{1.760 } & 4.723  \\
    Ours  &       & \cellcolor[rgb]{ .651,  .651,  .651}\textbf{7.337} & \cellcolor[rgb]{ .651,  .651,  .651}\textbf{61.301} & \cellcolor[rgb]{ .651,  .651,  .651}\textbf{15.905} & 1.744  & \cellcolor[rgb]{ .651,  .651,  .651}\textbf{4.850} \\
    \hline
    \end{tabular}%
  \label{tab:sota_fu_align}%
\end{table}%

Table~\ref{tab:sota_fu_mis} further evaluates the simulated uncalibrated setting.
Compared with C-OPDR, which estimates deformation fields for explicit cross-modal registration, BeyondFusion significantly improves all metrics under \(512\times512\) infrared inputs.
We also test BeyondFusion under the joint challenge of modality misalignment and reduced infrared resolution, with infrared inputs downsampled to \(128\times128\) and \(64\times64\).
The scores of BeyondFusion decrease only mildly as the infrared resolution is reduced, showing that BeyondFusion maintains stable fusion performance under simultaneous misalignment and infrared detail degradation.

\begin{table}[htbp]
  \centering
  \caption{Performance comparison of different methods on misaligned infrared-RGB image fusion. Gray cells indicate the best result.}
  \setlength{\tabcolsep}{3pt}
    \begin{tabular}{c|c|ccccc}
    \hline
    Methods & \diagbox{Resolution}{Metrics} & EN$\uparrow$  & SD$\uparrow$  & SF$\uparrow$   & SCD$\uparrow$ & AG$\uparrow$ \\
    \hline
    C-OPDR & \multirow{2}{*}{512$\times$512} & 6.888  & 46.127  & 9.900  & 1.335  & 3.322  \\
    Ours  &       & \cellcolor[rgb]{ .651,  .651,  .651}\textbf{7.316} & \cellcolor[rgb]{ .651,  .651,  .651}\textbf{60.089} & \cellcolor[rgb]{ .651,  .651,  .651}\textbf{16.673} & \cellcolor[rgb]{ .651,  .651,  .651}\textbf{1.659} & \cellcolor[rgb]{ .651,  .651,  .651}\textbf{5.099} \\
    \hline
    Ours  & 128$\times$128 & 7.321 & 59.797 & 16.258 & 1.653 & 4.992 \\
    \hline
    Ours  & 64$\times$64 & 7.301 & 59.444 & 15.673 & 1.640 & 4.751 \\
    \hline
    \end{tabular}%
  \label{tab:sota_fu_mis}%
\end{table}%

\begin{figure}[ht]
    \centering 
    \includegraphics[width=\columnwidth]{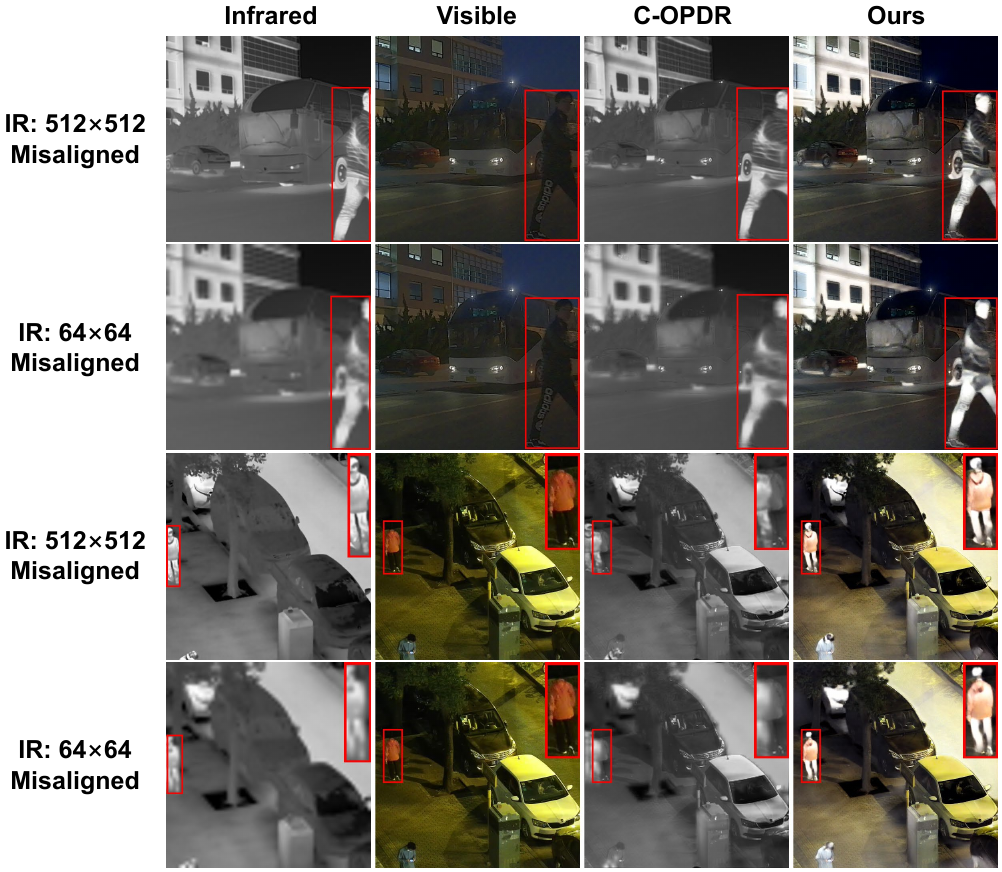}
    \caption{Qualitative comparison under simulated unregistered inputs (zoom in for details). BeyondFusion better preserves thermal saliency and visible structures under both spatial misalignment and infrared resolution degradation.}
    \label{fig:sota_fusion_mis}
\end{figure}

The visual comparison in Fig.~\ref{fig:sota_fusion_mis} further demonstrates the robustness of BeyondFusion.
Despite the spatial offsets between modalities, BeyondFusion integrates infrared details into the corresponding visible pedestrian regions in the fused image.
The fused pedestrians keep sharp body contours and clear local contrast without obvious shifted or duplicated artifacts.
By contrast, C-OPDR has difficulty maintaining consistent cross-modal correspondence under the challenging misalignment, so the fused pedestrian regions are shifted or partially detached from the visible human regions.
When the infrared input is further reduced to \(64\times64\), the fused pedestrian region becomes more blurred, while BeyondFusion still maintains clear pedestrian contours and well-localized infrared details.
These results indicate that BeyondFusion can maintain effective cross-modal interaction without explicit registration, making it more tolerant to spatial offsets and infrared resolution degradation.

\begin{figure}[ht]
    \centering 
    \includegraphics[width=\columnwidth]{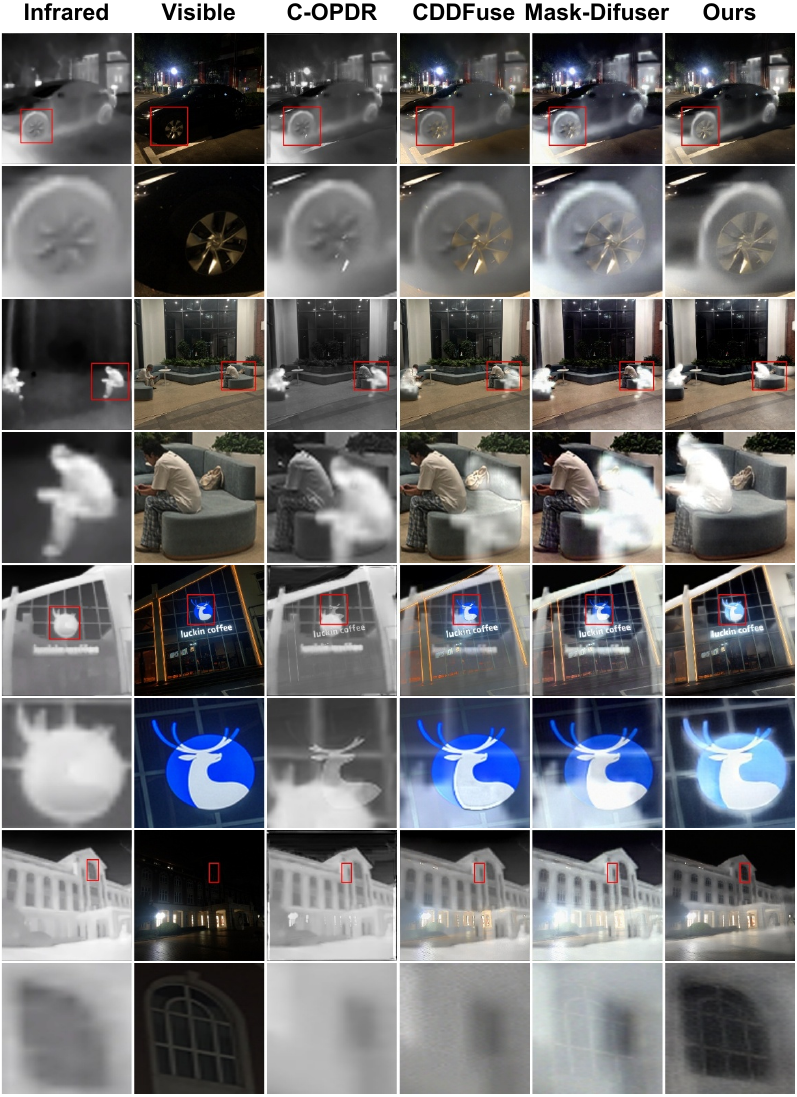}
    \caption{Qualitative infrared-visible image fusion comparison on our real-world smartphone dataset (zoom in for details). BeyondFusion shows strong generalization and robustness under misaligned infrared images.}
    \label{fig:fusion_val}
\end{figure}

Finally, Fig.~\ref{fig:fusion_val} evaluates fusion on the real mobile dataset.
The captured pairs include low-resolution infrared observations and high-resolution visible images.
Because the two mobile cameras have different viewpoints, camera parallax introduces natural spatial offsets between modalities, directly testing whether a fusion method can handle practical mobile acquisition.
CDDFuse and Mask-Difuser do not explicitly handle misalignment, and their outputs often behave like pixel-position fusion of mismatched infrared and visible contents.
C-OPDR introduces explicit registration, but still has difficulty handling such challenging real captures.
By contrast, BeyondFusion produces more coherent fused images: visible structures remain clear, and infrared details are incorporated into the corresponding object regions rather than being overlaid at mismatched pixel positions.
This advantage can be observed from the wheel hub, the sitting person, the Luckin Coffee logo, and the window railing in the building scene, where BeyondFusion better aligns infrared details with visible structures while suppressing obvious ghosting and misplaced highlights.

\subsection{Unified Training Analysis}
\label{train_pipeline}
BeyondFusion treats infrared super-resolution and infrared-visible fusion as two outputs of the same latent interaction process.
Before evaluating downstream perception, we examine whether this unified training strategy introduces a performance tradeoff compared with training two task-specific models separately.
Table~\ref{tab:train_pipeline} compares the two training strategies.

The performance gap between separate and joint training is small. 
Separate training results in slightly higher scores than joint training for both tasks.
These results indicate that sharing the latent denoising backbone does not introduce a substantial optimization conflict between the two outputs.
They also support the central premise of BeyondFusion: the two imaging objectives are compatible within the same latent generative process and can be optimized and executed together with acceptable performance sacrifice.
This makes a unified framework preferable to maintaining two isolated task-specific pipelines in this setting.

\begin{table}[htbp]
  \centering
  \caption{Comparison between separately trained task-specific models and a jointly trained unified model for infrared super-resolution and infrared-visible fusion.}
  \setlength{\tabcolsep}{3pt}
    \begin{tabular}{c|c|ccccc}
    \hline
    Task  & Metrics & PSNR$\uparrow$ & SSIM$\uparrow$ & LPIPS$\downarrow$ & MANIQA$\uparrow$ & MUSIQ$\uparrow$ \\
    \hline
    \multirow{2}{*}{Infrared SR} 
    & Separate & 30.38  & 0.8638 & 0.1658 & 0.4536 & 37.48 \\
    & Joint & 29.88 & 0.8570 & 0.1884 & 0.4230 & 36.28 \\
    \hline
    Task  & Metrics & EN$\uparrow$  & SD$\uparrow$  & SF$\uparrow$  & SCD$\uparrow$   & AG$\uparrow$ \\
    \hline
    \multirow{2}{*}{IVF}  & Separate & 7.301  & 59.444  & 15.673  & 1.640  & 4.751  \\
     & Joint & 7.320 & 59.092  & 15.321  & 1.615  & 4.625 \\
    \hline
    \end{tabular}%
  \label{tab:train_pipeline}%
\end{table}%

\subsection{Downstream Applications}
Beyond low-level image quality, whether the fused images benefit downstream perception is an important question for vision systems.
We therefore evaluate pedestrian detection on infrared-visible fusion results using the pretrained Grounded-SAM model~\cite{GSA} in a zero-shot setting.
All methods are evaluated with the same text prompt, \textit{person}, and the same detection pipeline.
Following the input assumptions of the compared fusion methods, CDDFuse and Mask-Difuser are evaluated on pixel-level aligned visible-infrared pairs with the infrared image downsampled to \(64\times64\), while C-OPDR is evaluated on misaligned visible-infrared pairs with the infrared image also downsampled to \(64\times64\).

\begin{figure}[ht]
    \centering 
    \includegraphics[width=\columnwidth]{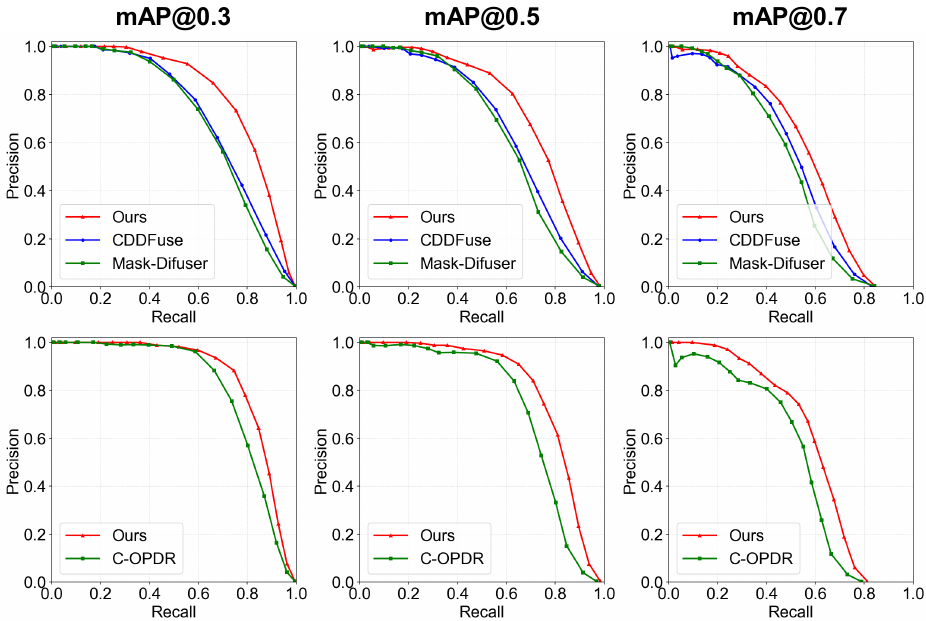}
    \caption{Precision-recall curve comparison for pedestrian detection with IoU thresholds of 0.3, 0.5, and 0.7.}
    \label{fig:mAP}
\end{figure}

Table~\ref{tab:mAP} reports the detection accuracy, and Fig.~\ref{fig:mAP} provides the corresponding precision-recall curves.
BeyondFusion consistently improves pedestrian detection across all IoU thresholds.
Compared with CDDFuse and Mask-Difuser in the aligned setting, BeyondFusion improves mAP by 0.053--0.108.
Compared with C-OPDR in the uncalibrated low-resolution setting, it improves mAP by 0.049--0.082.
The precision-recall curves show the same trend, with BeyondFusion maintaining higher precision over a wider recall range.
These improvements suggest that fewer fusion artifacts and clearer pedestrian structures in BeyondFusion outputs make them more suitable for downstream detection.

\begin{table}[htbp]
  \centering
  \caption{Pedestrian detection comparison with IoU thresholds of 0.3, 0.5, and 0.7.}
    \begin{tabular}{c|ccc}
    \hline
    \diagbox{Methods}{Metrics} & mAP@0.3 & mAP@0.5 & mAP@0.7 \\
    \hline
    CDDFuse & 0.7161  & 0.6577  & 0.5129  \\
    Mask-Difuser & 0.6990  & 0.6387  & 0.4904  \\
    Ours  & \cellcolor[rgb]{ .651,  .651,  .651}\textbf{0.8071 } & \cellcolor[rgb]{ .651,  .651,  .651}\textbf{0.7391 } & \cellcolor[rgb]{ .651,  .651,  .651}\textbf{0.5659 } \\
    \hline
    C-OPDR & 0.8029  & 0.7260  & 0.5065  \\
    Ours  & \cellcolor[rgb]{ .651,  .651,  .651}\textbf{0.8523 } & \cellcolor[rgb]{ .651,  .651,  .651}\textbf{0.8075 } & \cellcolor[rgb]{ .651,  .651,  .651}\textbf{0.5886 } \\
    \hline
    \end{tabular}%
  \label{tab:mAP}%
\end{table}%

\begin{figure}[ht]
    \centering 
    \includegraphics[width=\columnwidth]{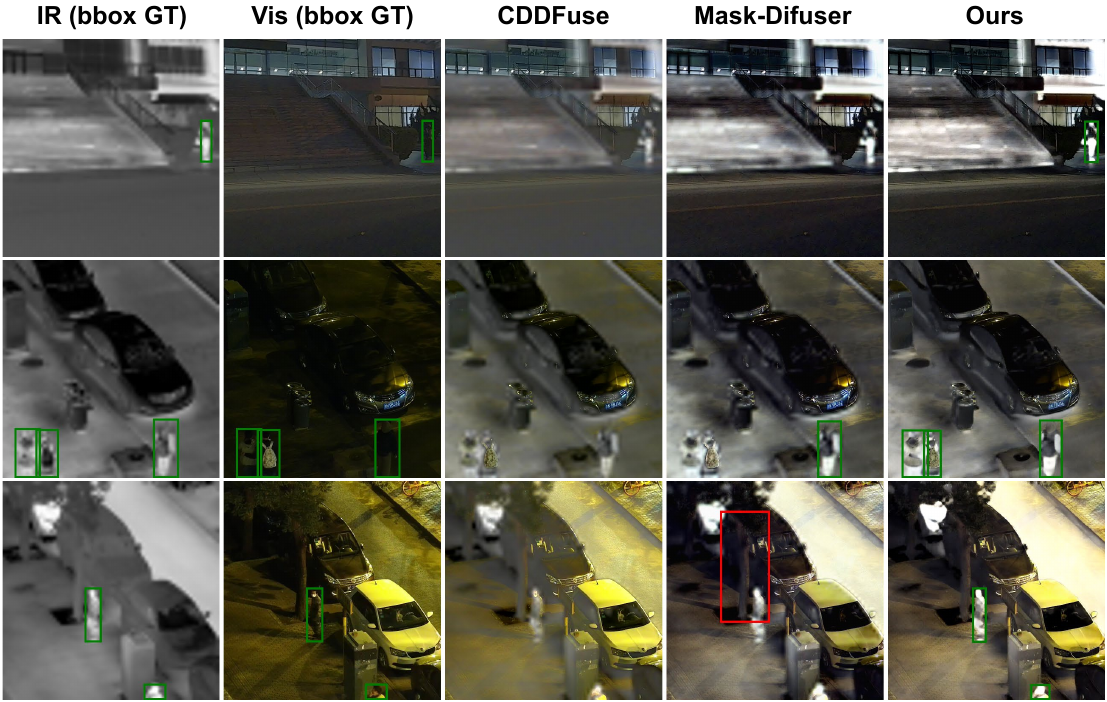}
    \caption{Visualization of comparison against CDDFuse and Mask-Difuser, where green bounding boxes indicate correct detections and red bounding boxes indicate incorrect detections.}
    \label{fig:cdd_detect}
\end{figure}

Fig.~\ref{fig:cdd_detect} further explains why the detection scores differ.
The lower detection performance of CDDFuse and Mask-Difuser is mainly caused by weakened pedestrian cues in their fused outputs.
In this figure, the ground-truth (GT) bounding boxes of pedestrians are overlaid on both input infrared and visible images for reference.
As shown in Rows 1--3, blurred pedestrian boundaries and weakened infrared saliency lead to missed or false detections in the baseline results.
In contrast, the fused images produced by BeyondFusion preserve clearer pedestrian semantics, enabling more accurate and reliable detection results.

\begin{figure}[ht]
    \centering 
    \includegraphics[width=\columnwidth]{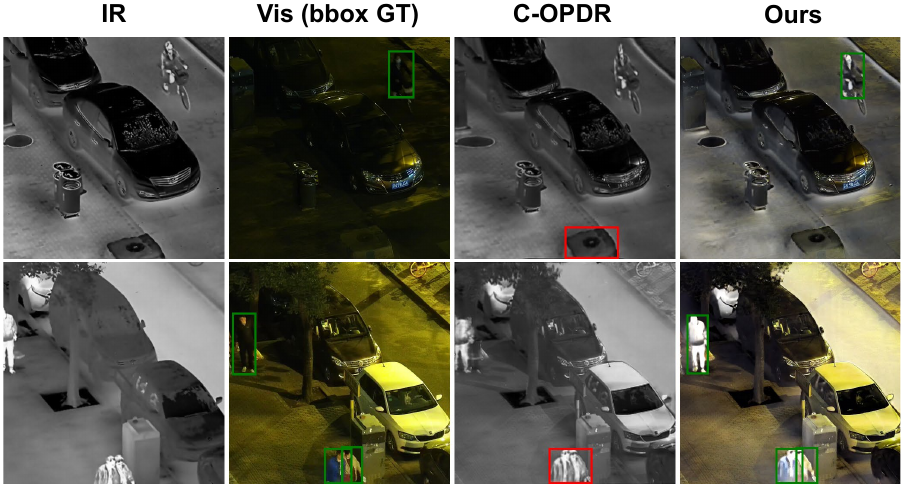}
    \caption{Visualization of comparison against C-OPDR, where green bounding boxes indicate correct detections and red bounding boxes indicate incorrect detections.}
    \label{fig:copdr_detect}
\end{figure}

The comparison against C-OPDR is shown in Fig.~\ref{fig:copdr_detect}. Since the input infrared images are spatially misaligned with the visible modality, the GT bounding boxes are overlaid only on the visible images for reference.
In Row 1, C-OPDR severely compresses the cyclist's body shape, making the detector fail to locate the true pedestrian region.
In Row 2, C-OPDR introduces a ghosting artifact around the reconstructed pedestrian in the upper-left region.
In contrast, BeyondFusion preserves a more coherent human posture and clearer pedestrian boundaries, enabling successful detection.

Overall, these results show that BeyondFusion remains robust to degraded or misaligned infrared inputs and generates semantically meaningful fused representations for downstream pedestrian detection, demonstrating its practical value beyond perceptual image quality.

\subsection{Ablation Study}
Table~\ref{tab:ablation} and Table~\ref{tab:ablation_fusion} each reports two groups of ablations with the joint training pipeline: removing the misalignment augmentation module (MA) from the full model and replacing CMSA with simpler interaction variants.
Removing MA degrades all metrics on infrared super-resolution and seriously deteriorates SCD score on infrared-visible image fusion, confirming the necessity of augmentation for uncalibrated visible-infrared inputs.
Replacing CMSA with the original self-attention (intra-modal attention only) or feature concatenation (without attention mechanism) also leads to weaker results on both tasks, indicating that simple intra-modal processing or static feature mixing does not provide the same latent workspace organization for robust cross-modal interaction.
With all components enabled, BeyondFusion achieves the best overall performance, supporting the effectiveness of the proposed unified design.

\begin{table}[htbp]
  \centering
  \footnotesize
  \caption{Ablation study of BeyondFusion components and interaction variants on infrared image super-resolution.}
  \setlength{\tabcolsep}{3pt}
    \begin{tabular}{c|ccccc}
    \hline
    \diagbox{Methods}{Metrics} & PSNR$\uparrow$ & SSIM$\uparrow$ & LPIPS$\downarrow$ & MANIQA$\uparrow$ & MUSIQ$\uparrow$ \\
    \hline
    w/o MA & 29.80 & 0.8537 & 0.2062 & 0.3788 & 32.85 \\
    \hline
    w/ Self-Attention & 29.99 & 0.8565 & 0.1960 & 0.4030 & 34.13 \\
    w/ Feature Concat & 29.50 & 0.8444 & 0.2386 & 0.3334 & 29.22 \\
    \hline
    w/ MA (CMSA) & 29.88 & 0.8570 & 0.1884 & 0.4230 & 36.28 \\
    \hline
    \end{tabular}
  \label{tab:ablation}
\end{table}

\begin{table}[htbp]
  \centering
  \footnotesize
  \caption{Ablation study of BeyondFusion components and interaction variants on infrared-visible image fusion.}
  \setlength{\tabcolsep}{3pt}
    \begin{tabular}{c|ccccc}
    \hline
    \diagbox{Methods}{Metrics} & EN$\uparrow$ & SD$\uparrow$ & SF$\uparrow$ & SCD$\uparrow$ & AG$\uparrow$ \\
    \hline
    w/o MA & 7.363 & 61.219 & 15.794 & 1.269 & 4.949 \\
    \hline
    w/ Self-Attention & 7.179 & 51.931 & 15.061 & 1.112 & 4.379 \\
    w/ Feature Concat & 7.317 & 54.542 & 15.192 & 1.406 & 4.453 \\
    \hline
    w/ MA (CMSA)  & 7.320 & 59.092 & 15.321 & 1.615 & 4.625 \\
    \hline
    \end{tabular}
  \label{tab:ablation_fusion}
\end{table}

%% file: sec/conclusion.tex
\section{Discussion and Conclusion}
BeyondFusion reflects a broader architectural direction for low-level vision: multimodal imaging can be approached as the organization of heterogeneous observations inside a shared generative computation, rather than as isolated network design for each degradation, modality, or output form.
In this work, infrared super-resolution and infrared-visible fusion are treated not as two independent pipelines, but as two readouts of one latent cross-modal process.
The same design principle can potentially support broader imaging problems that require joint use of multiple modalities, resolutions, frames, or degradation levels.
This flexibility is enabled by latent-token interaction.
By reorganizing self-attention tokens, BeyondFusion lets heterogeneous inputs exchange information inside the generative process without imposing fixed pixel-level alignment.
This is particularly useful for practical imaging systems, where observations often differ in modality, resolution, viewpoint, timing, and calibration state.
Such a framework provides a natural basis for future low-level vision systems that must operate under realistic acquisition constraints.

In conclusion, we presented BeyondFusion, a self-aligned latent diffusion framework for calibration-free infrared super-resolution and infrared-visible image fusion.
By reorganizing visible and infrared latent tokens within the self-attention layers of a pretrained diffusion model, BeyondFusion enables implicit cross-modal interaction without explicit registration.
Experiments on public benchmarks and a real mobile visible-infrared imaging system demonstrate its effectiveness under aligned, misaligned, low-resolution, and practical mobile settings, as well as its benefit for downstream pedestrian detection.